\documentclass[letterpaper, 10 pt, conference]{ieeeconf}  

\IEEEoverridecommandlockouts                              

\let\labelindent\relax

\usepackage[table,xcdraw]{xcolor}
\usepackage{graphics} 
\usepackage{epsfig} 
\usepackage{mathptmx} 
\usepackage{times} 
\usepackage{amssymb}  
\usepackage{amsmath} 
\usepackage{placeins}
\usepackage{enumitem}
\usepackage{caption}
\usepackage{subcaption}
\usepackage[ruled,linesnumbered]{algorithm2e}

\newtheorem{remark}{Remark}
\newtheorem{lemma}{Lemma}
\newtheorem{definition}{Definition}
\newtheorem{theorem}{Theorem}
\title{\LARGE \bf
Multi-agent Collective Construction using 3D Decomposition
}

\author{Akshaya Kesarimangalam Srinivasan$^{1}$, Shambhavi Singh$^{2}$, Geordan Gutow$^{3}$, Howie Choset$^{4}$ and Bhaskar Vundurthy$^{5}$
\thanks{*This work was not supported by any organization}
\thanks{$^{1}$Akshaya Kesarimangalam Srinivasan is a Masters student in the Robotics Institute at Carnegie Mellon University, USA}%
\thanks{$^{2}$Shambhavi Singh is an intern in the Robotics Institute at Carnegie Mellon University, USA, and a student at Birla Institute of Technology and Science, Pilani, India}%
\thanks{$^{3}$Geordan Gutow is a Postdoctoral fellow in the Robotics Institute at Carnegie Mellon University, USA}%
\thanks{$^{4}$Howie Choset is a Professor with the Robotics Institute at Carnegie Mellon University, USA}%
\thanks{$^{5}$Bhaskar Vundurthy is a Postdoctoral fellow in the Robotics Institute at Carnegie Mellon University, USA}%
}

\begin{document}

\maketitle
\thispagestyle{empty}
\pagestyle{empty}

\begin{abstract}

This paper addresses a Multi-Agent Collective Construction (MACC) problem that aims to build a three-dimensional structure comprised of cubic blocks. We use cube-shaped robots that can carry one cubic block at a time, and move forward, reverse, left, and right to an adjacent cell of the same height or climb up and down one cube height. To construct structures taller than one cube, the robots must build supporting stairs made of blocks and remove the stairs once the structure is built. Conventional techniques solve for the entire structure at once and quickly become intractable for larger workspaces and complex structures, especially in a multi-agent setting. To this end, we present a decomposition algorithm that computes valid substructures based on intrinsic structural dependencies. We use Mixed Integer Linear Programming (MILP) to solve for each of these substructures and then aggregate the solutions to construct the entire structure. 

Extensive testing on 200 randomly generated structures shows an order of magnitude improvement in the solution computation time compared to an MILP approach without decomposition. Additionally, compared to Reinforcement Learning (RL) based and heuristics-based approaches drawn from the literature, our solution indicates orders of magnitude improvement in the number of pick-up and drop-off actions required to construct a structure. Furthermore, we leverage the independence between substructures to detect which substructures can be built in parallel. With this parallelization technique, we illustrate a further improvement in the number of time steps required to complete building the structure. This work is a step towards applying multi-agent collective construction for real-world structures by significantly reducing solution computation time with a bounded increase in the number of time steps required to build the structure.

\end{abstract}

\section{INTRODUCTION}

There is a growing class of applications in which robots are used to assemble and construct structures \cite{quadrotor_teams}. Some robotics technologies for on-site building construction \cite{on_site_construction} include additive manufacturing \cite{additive_manufacturing}, automated robotic assembly \cite{steel_beam}, and bricklaying \cite{brick_laying}. It is particularly relevant in settings like open pit mining and extraterrestrial or underwater construction where human presence is difficult or dangerous \cite{stigmergy}\cite{open_pit_mining}\cite{extra_terrestrial}\cite{weightless_construction}. Delegating construction to robots in these scenarios can thus be beneficial. Automation can also improve construction speed, and efficiency \cite{TERMES}. In these applications, teams of smaller robots are often more effective than a few larger robots as they are cheaper, easier to deploy, and facilitate more parallelization \cite{tree_approach}\cite{quadrotor_teams}\cite{multi_material_construction}.


\begin{figure}[h!]
        \centering
        \begin{subfigure}[b]{0.475\columnwidth}
            \centering
            \includegraphics[width=\linewidth]{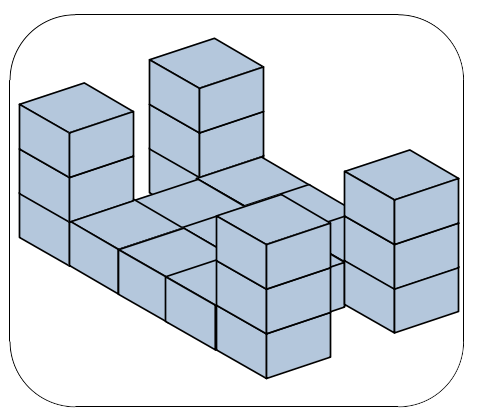}
            \caption[]%
            {{\small Input structure}}    
            \label{fig:1_a}
        \end{subfigure}
        \begin{subfigure}[b]{0.475\columnwidth}  
            \centering 
            \includegraphics[width=\linewidth]{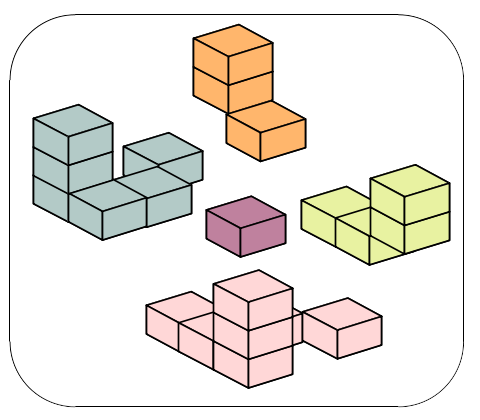}
            \caption[]%
            {{\small Structural decomposition}}    
            \label{fig:1_b}
        \end{subfigure}
        \vskip\baselineskip
        \begin{subfigure}[b]{0.475\columnwidth}   
            \centering 
            \includegraphics[width=\linewidth]{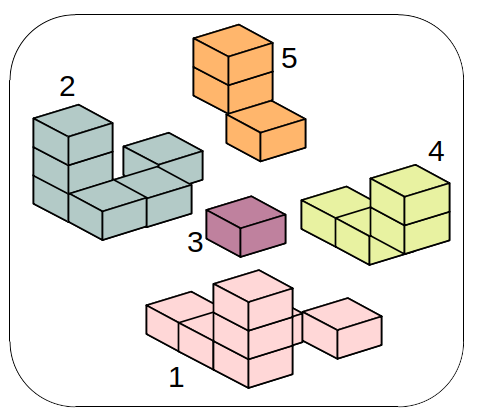}
            \caption[]%
            {{\small Ordering of substructures}}    
            \label{fig:1_c}
        \end{subfigure}
        \begin{subfigure}[b]{0.475\columnwidth}   
            \centering 
            \includegraphics[width=\linewidth]{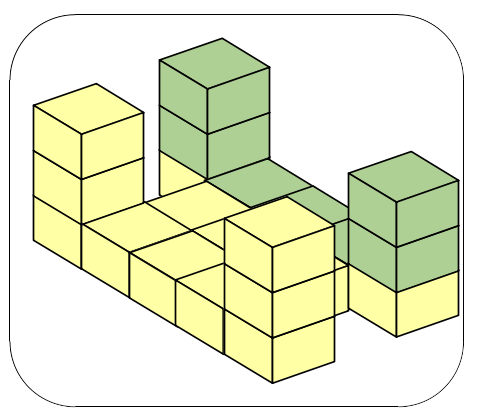}
            \caption[]%
            {{\small Parallelizable construction}}    
            \label{fig:1_d}
        \end{subfigure}
        \caption[]
        {\small Visualization of our solution to the MACC problem.}
        \label{fig:motivating_image}
    \end{figure}

The multi-agent collective construction (MACC) problem aims to construct a given three-dimensional structure using a set of robots in a grid world in the minimum number of time steps \cite{EA}. Previously the MACC problem has been solved using optimization \cite{EA}, heuristics \cite{tree_approach}\cite{control_rules_construction}, \cite{temporal_goal_sequencing}, or Reinforcement Learning (RL) \cite{RL}, \cite{RL_in_dynamic_env}. All of these approaches have severe limitations. Although \cite{EA} finds an optimal solution, it has a very high solution computation time, even for some simple structures. \cite{tree_approach} requires less computation time but provides high-cost solutions. While the approach taken in \cite{RL} is cheap to evaluate on a novel structure, it requires large amounts of training data and does not reliably produce the desired structure.
 
We present our work on decomposing an input structure into substructures and using an existing Mixed Integer Linear Programming (MILP) formulation \cite{EA} for each substructure to find good-quality solutions in a reasonable amount of time. This is conceptually visualized in Fig. \ref{fig:motivating_image}. 

Our main contributions are as follows:
\begin{enumerate}
    \item An algorithm to decompose an input structure into substructures whose construction may be planned independently
    \item An algorithm to find an order in which the substructures can be built
    \item An approach to leverage dependency between substructures to identify those that can be built simultaneously
    \item Extensive numerical results demonstrating the computational improvements over existing methods 
\end{enumerate}

The problem formulation is presented in section \ref{problem_formulation}. Then, existing approaches to the multi-agent collective construction problem are outlined in section \ref{related_work}. Section \ref{approach} presents the decomposition and bottom-up planning algorithms proposed to solve this problem. The extension of the algorithm to parallel construction is described in section \ref{parallel_construction}. Section \ref{experiments} provides numerical results demonstrating the approach's effectiveness. Section \ref{conclusion} presents some conclusions and possible directions for future work. 

\section{PROBLEM FORMULATION}\label{problem_formulation}

Similar to \cite{EA}, the MACC problem in this work is set in a 3D grid world with two principal components: a predefined structure comprised of cube-shaped building \textit{blocks} and block-sized \textit{robots}. The robots collaboratively construct the structure by moving these blocks using any of the following permissible actions in a given time step: 
\begin{enumerate}
    \item Move to a free cell in the four compass directions 
    \item Climb up or down one block height to an adjacent occupied cell
    \item Pick up or place a block at an adjacent cell of the same height
\end{enumerate}

The standard rules of gravity apply to all the blocks, and it is assumed that robots always interact with the topmost blocks. In order to access a cell not surrounded by blocks of the same height, the robots construct scaffolding. All scaffolding must be removed after the completion of the entire structure. We further assume an unlimited supply of blocks at the boundary of the grid world and disregard any movement of the robots outside the grid world. In other words, the robots that exit the grid world can enter from any boundary cell in the next time step. The world is initially empty, and the robots start and finish outside the grid world.

This work discusses time-efficient algorithms to obtain a sequence of actions for $N$ robots to collectively construct a predefined structure. We present a comparison of our solutions with state-of-the-art via the following metrics: 
\begin{itemize}
    \item \textbf{Computation Time:} Time taken to compute the action sequence for building the structure
    \item \textbf{Makespan:} Total number of time steps to complete building the structure
    \item \textbf{Sum of costs:} Total number of actions for $N$ robots to build the structure
\end{itemize}

\section{RELATED WORKS}\label{related_work}

Multi-agent collective construction has been explored in both two \cite{2D_construction} and three-dimensional worlds with varying types of agents and construction blocks. For instance, \cite{quadrotor_teams} uses a team of quadrotors to build structures made of beams and columns. Harvard's TERMES project, on the other hand, addresses the problem with homogeneous blocks and agents \cite{TERMES}. They show how teams of smaller robots are effective at collectively building structures much larger than themselves. The TERMES project inspired several works that proposed ways to find a sequence of actions for TERMES-like robots to build structures.

\cite{tree_approach} solves the problem by performing dynamic programming on a minimum spanning tree that spans the cells of the weighted workspace and restricts the agent's movements to the edges of this tree. This planning method minimizes the number of pick-up and drop-off operations but is restricted to the single agent case. \cite{local_tree} extends this to multiple agents by parallelizing the action sequence. It performs a local search to find a more cost-efficient spanning tree using a recursive algorithm. However, this increases the algorithm's complexity, and the solution is still not optimal. 

The distributed multi-agent reinforcement learning algorithm used in \cite{RL} extends single-agent advantage actor-critic to enable multiple agents to learn a homogeneous, distributed policy. The learned policy is tested with swarms of various sizes on structures not seen during the nine days of training. Although this approach is complete, some structures were not built accurately. Moreover, the makespan of the test structures was 100 times more than the minimum possible makespan found by optimization approaches.

The work most closely related to the current effort is the optimization approach presented in \cite{EA}. The MACC problem is solved using MILP or Constraint Programming (CP). The MILP model treats all robots as one flow through a time-expanded graph. Each decision variable defines the action the robot took and if it was carrying a block at that time step. The dynamics of the world are modeled as constraints for the set of decision variables. The formulation also models each cell in the workspace as pillars that aim to reach their target height. The CP approach uses a simpler network flow structure than the MILP approach. CP models the specifications of the world, such as the vertical dimension, actions, and block-carrying state, as logical and element constraints that better exploit the strength of CP. Both formulations use an objective function that minimizes the sum of costs and externally minimizes makespan. The MILP model computed the globally optimal makespan and the optimal sum of costs for a particular makespan. Both optimization models found feasible solutions to the six test instances with makespans less than 20. However, the MILP formulation required less computation time than the CP formulation. The approach presented in Section \ref{approach} calls this MILP formulation as a subroutine.  

The low makespans of the structures studied in \cite{EA} indicate that small structures are simple to construct as they do not need large makespans. However, even for some of the six small structures, the optimization models needed more than five days to compute a solution. As will be further demonstrated in section \ref{experiments}, the MILP solution computation time is excessive for complex structures requiring longer makespans to complete construction. This emphasizes the need to find approaches with a practical solution computation time for structures of varying complexity.

There has been some work, including \cite{scalable_MILP}, \cite{large_MILP} for solving these large MILP problems. \cite{large_MILP} uses a two-level approach to make large MILP problems tractable. It first coarsens binary variables to reduce the number of variables in the MILP problem and forms a semi-coarse model. It then aggregates constraints by partitioning them into groups and adding the violated constraints to the semi-coarse model iteratively till all the constraints in the full model are added. Inspired by 3D model decomposition work \cite{3D_decomposition}, this paper reduces the number of variables in the MILP problem by solving for one substructure at a time. Constraints are aggregated at each stage to represent the intermediate structure to be built.

\section{APPROACH}\label{approach}

In this section, we propose a decomposition algorithm to break down predefined structures into simpler substructures. We then determine an order in which it is possible to build the substructures and utilize Mixed Integer Linear Programming (MILP) to compute an optimal sequence of actions for every substructure \cite{EA}.





\subsection{Structural Decomposition} 
In a 3D grid world of dimensions $(X\times Y \times Z)$, we denote predefined structures using $\bar{z}(x,y)$ where $\bar{z}$ indicates the height of the topmost block at every grid location $(x,y)$, $x,y,z,\bar{z} \geq 0$ and $x\leq X, y\leq Y, z\leq \bar{z}\leq Z$. We begin by ensuring that all predefined structures are valid and performing a similar check for substructures. 

\begin{definition}
\label{def:valid}
A structure is considered to be \textbf{valid} if, for every block of the structure at height $z$ $(z>1)$, there exists a block at height $(z-1)$. 
\end{definition}

Definition \ref{def:valid} ensures that any movement in the blocks is achieved only through the robots' permissible actions.  A valid structure always has a sequence of permissible actions that constructs the structure. It is worth noting that the robots can utilize blocks at a lower height as scaffolding for higher blocks. Such an action minimizes duplication of efforts from building temporary scaffolding for every block in the structure. As a result, it is preferable to ensure that any useful blocks (for scaffolding) are part of the same substructure as the higher block under consideration. We use the notion of a \textit{shadow region} to identify the relation between neighboring blocks and identify substructures. 


\begin{definition}
For a tower of height $z$ located at $(x,y)$, all cells $(x',y',z')$ s.t. \begin{equation*}|x-x'| + |y-y'| < h\end{equation*} and \begin{equation*}z' \leq h - (|x-x'|+|y-y'|)\end{equation*} are part of the \textbf{shadow region} of the tower. 
\end{definition}

Substructures are inherently associated with a specific order in that the validity of a substructure depends on which substructures are already present. 

\begin{definition}
Consider a set of substructures represented by $S_1$ to $S_d$ where $'d'$ is the number of substructures. Let $\bigcup_{i=1,2,\cdots,j-1}S_i$ be a valid structure. Then $S_j$ is a \textbf{valid substructure} if $\bigcup_{i=1,2,\cdots,j} S_i$ is also a valid structure.
\end{definition}

In the context of our problem, we define the basin of attraction of a tower of height $h$ in the structure to be all cells in the shadow region of that tower. Equivalently, all the structure blocks that can help build a tower by acting as direct support or scaffolding are part of its basin of attraction.

The decomposition algorithm iterates through the towers in the input structure in decreasing order of height. At each step, the blocks in the shadow region of the tower that are not already part of another substructure become part of its substructure. This decomposition algorithm is described in Algorithm \ref{alg:basins_of_attraction}.

\begin{algorithm}
  ($X$,$Y$,$Z$) $\gets$ grid world dimension;
  \\
  $\bar{z}$ $\gets$ input structure;
  \\
  $towers$ $\gets$ non-zero $\bar{z}$ elements in decreasing order;
  \\
  Initialize the set of all substructures, $sub$ = $\varnothing$;
  \\
  \For{$h$ in $towers$}{
    \If{Topmost block of tower $h$ already in a substructure}{
        continue;
    }
    Initialize substructure of $h, sub_h$ $\gets$ $\varnothing$;
    \\
    Shadow indices, $s_{idx}$ $\gets$ grid cells in shadow region of tower $h$;
    \\
    \For{$s$ in $s_{idx}$}{
        \If{$s$ not in any substructure}{
            add $s$ to $sub_h$;
        }
    }
    Add $sub_h$ to $sub$;
  }
  \caption{Decomposition using basins of attraction}\label{alg:basins_of_attraction}
\end{algorithm}

\begin{theorem}
Each substructure in the order they are found by algorithm \ref{alg:basins_of_attraction} is a valid substructure.
\label{theorem1}
\end{theorem}

Proof of Theorem \ref{theorem1} is omitted in this paper for brevity. 

\subsection{Bottom Up Planning}

\begin{figure}
  \centering
  \includegraphics[width=0.7\linewidth]{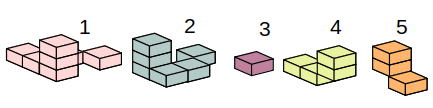}
  \caption{Substructures identified by Algorithm \ref{alg:basins_of_attraction}}
  \label{fig:feasibleorder1}
\end{figure}

Consider the example shown in Fig. \ref{fig:motivating_image}. Algorithm \ref{alg:basins_of_attraction} identifies five substructures (see Fig. \ref{fig:feasibleorder1}), and hence there are 120 different possible orders in which the substructures can be constructed. However, building all substructures will not be possible in some of those orders. For example, consider building the substructures in the order of (2),(4),(5),(1), and (3). This is not possible as (4) requires (1) to be built before it, and (3) cannot be built once (1), (2), (4), and (5) are constructed. Hence, the order of construction of the substructures is not trivial.

To find an order of construction of substructures, i.e., a sequence of substructures, we define the traversability property of substructures. As with validity, this is depends on which other substructures are already present.


\begin{definition}
A substructure is \textbf{traversable} if, for every block in the substructure, there exists a feasible path from the boundary to at least one neighbor cell at the same height as the block in the presence of previously constructed substructures.
\end{definition}

\begin{definition}
A \textbf{feasible path} is a sequence of permissible actions, not including picking up blocks from previously constructed substructures.
\end{definition}

\begin{lemma}
If there is no neighbor cell at the same height as a block to be placed in a substructure, as a consequence of Algorithm \ref{alg:basins_of_attraction}, there will always be enough space to build a scaffolding of the required height.
\label{lemma:scaffolding_space}
\end{lemma}

The proof of Lemma \ref{lemma:scaffolding_space} is omitted in this paper for brevity. 

\begin{definition}
A sequence of substructures is \textbf{feasible} if $\forall i \in {1,\cdots,d}$, $S_i$ is traversable and $\bigcup_{k=1,2,\cdots,i}S_k$ is a valid structure.
\end{definition}

Obtaining a feasible sequence of substructures is treated as an assembly sequencing problem where the substructures are the components, and the input structure is the final product. The key idea is to build substructures in the reverse order in which they can be removed/disassembled from the goal structure while leaving at every step a valid structure. Such an order is obtained via a bottom-up planning algorithm inspired from \cite{assembly_sequencing_bottom_up_planning}, \cite{structure_oriented_assembly_sequencing}, presented in Algorithm \ref{alg:bottom_up_planning}. 

\begin{remark}
Henceforth, the index of a substructure refers to the order in which the decomposition algorithm found the substructures.
\end{remark}

To find the order in which substructures can be removed, we need to check if each block in a substructure is removable. From the problem formulation:
\begin{enumerate}
    \item To remove a block at $(x,y,z)$, there should be no block at $(x,y,z+1)$
    \item To pick up a block at $(x,y,z)$, the robot needs to be at a neighbor cell of $(x,y)$ at the same height $z$ 
    \item If there are no neighbor cells at height $z$, it needs to have enough space to build scaffolding to height $z$\label{cond_pt3}
\end{enumerate}

These conditions can be reduced further. Condition \ref{cond_pt3} is always satisfied by algorithm \ref{alg:basins_of_attraction} as stated in Lemma \ref{lemma:scaffolding_space}.

Thus, the necessary and sufficient conditions to check if a substructure is removable can be reduced to two:
\begin{enumerate}[label=(\alph*)]
    \item No block in the substructure should have a block from another substructure on top of it
    \item There is a traversable path for a robot from outside the grid to at least one of the neighboring cells for each of the blocks in the substructure, i.e., the substructure should be traversable in the current state of the environment
\end{enumerate}

Condition one implies that the only blocks allowed to be on top of a block $b$ in substructure $S$ are blocks that are part of $S$ too. Thus, the top blocks will be removed while removing $S$, ensuring block $b$ is removable. This yields a notion of dependence between substructures:

\begin{definition}
A substructure $S_i$ is said to be \textbf{dependent} on substructure $S_j$, (denoted $S_i \xrightarrow[]{} S_j$) if $\bigcup_{\substack{k=1,2,\cdots,i\\ k \neq j}} S_k$ is not a valid substructure.
\end{definition}

To check condition 2, a traversability matrix is constructed at every stage, considering all the substructures yet to be removed. The $(i,j)^{th}$ element is 1 only if the $(i,j)$ location is reachable from outside the grid. This is determined using dynamic programming starting from the boundary cells (which are always reachable). Every subsequent cell is reachable if a neighboring cell is reachable and the neighbor has a height difference of less than two from the cell (as the robots can only climb or descend one cube height at a time). Once this matrix is obtained, contour polygons of unreachable cells are computed. These contours represent impassable walls in the environment. Blocks are treated as not removable if they are enclosed on all four sides by blocks from another substructure or inside a contour of impassable walls. Note that in certain cases (the presence of a staircase on the wall's interior), scaffolding would, in principle, allow passing these walls. Thus this check is a sufficient but not necessary condition for removability. This algorithm is described in Algorithm \ref{alg:bottom_up_planning_removable}.

Finally, a feasible sequence of substructures can be generated. Let $O_d$ be the order in which the substructures were found and $O_{new}$ be the new feasible ordering. Algorithm \ref{alg:bottom_up_planning} iterates through $O_d$ in reverse order and adds only substructures that can be removed using Algorithm \ref{alg:bottom_up_planning_removable} to $O_{new}$. During one pass through $O_d$, if none of the substructures were removable, this implies that a substructure $S_i$ was not traversable due to substructure $S_j$ and $S_i \xrightarrow[]{} S_j$. In this rare scenario, the two substructures are merged, resulting in a traversable substructure. This is repeated until all substructures are added to $O_{new}$. The final feasible sequence is computed as the reverse of $O_{new}$.

\begin{remark}
None of the 206 different structures considered in section \ref{experiments} required merging to obtain a feasible sequence of substructures.
\end{remark}

\begin{lemma}
The reverse of the order of removing substructures is a feasible assembly order. 
\label{lemma:reverse_order}
\end{lemma}

The proof of Lemma \ref{lemma:reverse_order} is omitted in this paper for brevityx.

\begin{algorithm}
$S$ $\gets$ substructure being checked
\\
$traversability\_matrix$ $\gets$ reachable positions in the (x,y) grid
\\
$contours$ $\gets$ polygons representing impassable enclosures in the traversability matrix
\\
\For{$b$ $\in$ blocks of $S$}{
    \If{$b$ surrounded by blocks from another substructure in all four directions}{
        \KwRet False;
    }
    \If{$b$ inside any contour $\in$ $contours$}{
        \KwRet False;
    }
}
\KwRet True;
\caption{substructure removable check}\label{alg:bottom_up_planning_removable}
\end{algorithm}

\begin{algorithm}[h!]
$O_d$ $\gets$ original order;
\\
$O_{new}$ $\gets$ feasible order;
\\
\While{all substructures are not ordered}{
    \For{$sub$ in reverse order in $O_d$}{
        \If{$sub$ in $O_{new}$}{
            continue;
        }
        \If{removable($sub$)}{
            delete $sub$ from $O_d$;
            \\
            append $sub$ to $O_{new}$;
        }
    }
    \If{no $sub$ was removed and all substructures not ordered}{
        Merge last two substructures in $O_d$ that are not in $O_{new}$
    }
}
\KwRet reverse of $O_{new}$
\caption{substructure ordering}\label{alg:bottom_up_planning}
\end{algorithm}

\begin{theorem}
Given valid substructures, algorithm \ref{alg:bottom_up_planning} always generates a feasible sequence of substructures. 
\label{theorem2}
\end{theorem}

\begin{lemma}
Given a valid input structure, the decomposition and ordering algorithms followed by MILP optimization for each substructure return a legal/feasible action sequence to build the structure if one exists. 
\label{lemma:MILP}
\end{lemma}

\begin{theorem}
Consider a structure decomposed into $d$ substructures. The number of time steps required to construct the structure by constructing substructures sequentially is no more than $d$ times the number of time steps required to construct the structure without decomposition.
\label{theorem3}
\end{theorem}

Proofs of Theorems \ref{theorem2}-\ref{theorem3} and Lemma \ref{lemma:MILP} are omitted in this paper for brevity.

\subsection{Parallel Construction}
\label{parallel_construction}
Once the substructures and a feasible order are obtained, any of the conventional approaches (\cite{EA}, \cite{RL} and \cite{tree_approach}) can be used to find the sequence of actions to build each substructure.  However, if mixed integer linear programming as in \cite{EA} is used, some substructures can be built in parallel. This reduces the total time to build the structure. 
\\
This is achieved by modifying the bottom-up planner. In sequential construction of substructures: any input is decomposed using Algorithm \ref{alg:basins_of_attraction} into substructures ordered using Algorithm \ref{alg:bottom_up_planning}. For parallel construction, Algorithm \ref{alg:bottom_up_planning} is modified such that at every stage, all the substructures that can be removed are determined instead of just the substructure being considered as per $O_{d}$ (default order). These substructures can potentially be built in parallel. 
\\
Let $SP$ be a set of substructures that can be built in parallel. The first structure in $SP$ is built as described for sequential construction. For every subsequent substructure $S_i$ in $SP$, the actions taken to build all previous $S_j$ for $j \in {1,\cdots,i-1}$ are added as a constraint to the MILP. This ensures that $S_i$'s solution avoids agent-agent collision with the agents building the previous substructure and does not use more agents than permitted. In this approach, in the worst case, if the substructures in $SP$ cannot be built in parallel (due to, say, insufficient agents), the ordering algorithm will provide a sequential order.
\\
Parallel execution reduces makespan at the cost of increased constraints in the MILP formulations for each substructure. However, our initial experiments show that this does not significantly affect the solution computation time.

\section{RESULTS}
\label{experiments}

\subsection{Experimental Setup}
We evaluate the effectiveness of our algorithm on a variety of test structures: (a) six test cases used by \cite{EA} and \cite{RL}, (b) one hundred randomly generated structures in a 10x10x4 grid world, and (c) one hundred randomly generated structures in a 7x7x4 grid world.

The MILP approach obtains the action sequence to build each structure or substructure per the experiment. All optimization models are solved using Gurobi 9.0.2, a state-of-the-art solver for MILP on an Intel® Core™ i7-7700K CPU @ 4.20GHz × 8 with 94GB of memory. In every case, the Gurobi model was allowed to run for up to 10,000 seconds for each structure. 

\subsection{Results on an Example Structure}

We first demonstrate our method on the example structure from Fig. \ref{fig:motivating_image}. As shown in Fig. \ref{fig:motivating_image} (b), the algorithm gives five substructures. The MILP model gets each substructure as an input and gives a set of action sequences with a maximum of 20 robots for each substructure. Table \ref{tab:reference structure} presents the metrics of the solution for each substructure. The approach in \cite{EA} iterates through increasing makespans until the model becomes feasible and subsequently performs optimization to find a solution. `Solve Time' denotes the time taken to optimize the final model, and `Total Solve Time' denotes the time to iterate through all time steps along with the solve time of the model.
\begin{table}[!htb]
\setlength\tabcolsep{0pt}
\caption{Results on the structure used in Fig. \ref{fig:motivating_image} illustrating the metrics for each substructure indicated by Sub. No.} \label{tab:reference structure}
\centering
\smallskip
\begin{tabular*}{\columnwidth}{@{\extracolsep{\fill}}rccccl}
  Sub. No. &Makespan & Sum-of-costs & Solve Time & Total Solve Time\\
    \hline
1 &  14   & 68     & 37.0s & 123.6s\\
2&  14   & 50      & 8.6s & 96.8s\\
3&  14   & 42      & 7.4s & 95.5s\\
4&  15   & 39      & 6.7s & 111.4s\\
5&  10   & 8     & 1.3s & 34.6s\\
  \hline
Total & 67   & 207     & 61.0s &461.9s\\
\end{tabular*}
\end{table}

\subsection{Comparison with Other Approaches on Six Test Structures}

We next conduct experiments to compare our approach with two existing sub-optimal approaches \cite{tree_approach}\cite{RL}. The experiments use a set of six test structures also used by \cite{EA} and \cite{RL} for reporting their results. The Reinforcement Learning approach \cite{RL} runs a pretrained policy for 100 trials using eight agents for each structure. The authors noted that using more than eight agents detriments the performance. 



Table \ref{tab:Baseline comparison} reports the sum of costs for the tree approach \cite{tree_approach} and the average sum of costs for the RL approach (over successful trials). It also reports the sum of costs using the MILP approach with decomposition (our approach). Since these existing methods aim to minimize the number of pick-up and drop-off actions alone, our approach though not optimal achieves costs nearly an order of magnitude smaller than these methods.

\begin{table}[h!]
\setlength\tabcolsep{10pt}
\caption{Comparison of the sum of costs of solutions for our approach with other non-optimal methods} \label{tab:Baseline comparison}
\begin{tabular}{rccccl}
Structure & Tree based \cite{tree_approach} & RL based \cite{RL}  & Ours\\ \hline
1         & 1144                & 3040                   & 179\\
2         & 836                 & 1026                    & 128\\
3         & 1590                & 3056                   & 326\\
4         & 2120                & 3252                   & 263\\
5         & 2180                & 2804                 & 381\\
6         & 808                 & 1276                   & 161\\
\end{tabular}
\end{table}

Finally, we compare our approach with an existing optimal approach \cite{EA}. \cite{EA} uses the entire structure as input
to their MILP formulation. In our experiments, the number of robots is limited to 20 for uniformity between solving with and without decomposition. Table \ref{tab:MILP comparison} presents results for the approach from \cite{EA} and our decomposition technique with and without parallelism. Decomposition significantly improves the solution computation times over pure MILP while maintaining a similar sum of costs. However, we observe an increase in time steps when constructing the structure via substructures. Introducing parallel construction largely mitigates this increase. On average, parallel construction reduces the number of time steps by 46\% compared to pure decomposition with a similar sum of costs and solution computation time. Note that, for structures 4-6, parallel construction significantly increases the solution computation time. However, it is still much faster than MILP without decomposition.

\begin{figure}[h!]
    \centering
    \includegraphics[width=0.75\linewidth]{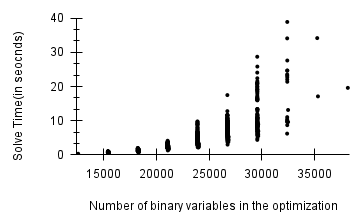}
    \centering
    \caption{The total computation time when using MILP to solve for the entire structure is exponential in the number of optimization variables.}
    \label{fig:solve time variables}
\end{figure}

\begin{table*}[h!]
\setlength\tabcolsep{10pt}
\caption{Results for the six test cases} \label{tab:MILP comparison}
\centering
This table presents a comparison of metrics for solutions obtained using our approach with the time-optimal solution presented in \cite{EA} on a set of six test structures used by \cite{EA} and \cite{RL}. Numbers indicated in {\color[HTML]{32CB00} green} show the improvements of our approach compared to the state of the art. Here, \textbf{A - MILP-based Exact Approach, B - 3D decomposition with MILP-based Approach and C - Parallel Construction of substructures using 3D decomposition with MILP-based Approach}
\begin{tabular}{cccccccccc}
\hline
\multicolumn{1}{|c|}{Test Structure}& \multicolumn{3}{c|}{\includegraphics[width=0.15\linewidth]{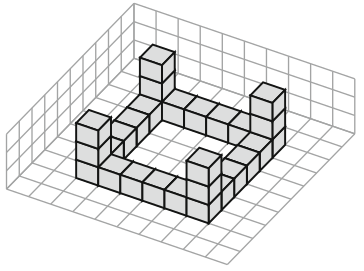}} & \multicolumn{3}{c|}{\includegraphics[width=0.15\linewidth]{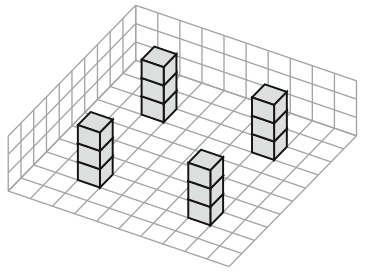}} & \multicolumn{3}{c|}{\includegraphics[width=0.15\linewidth]{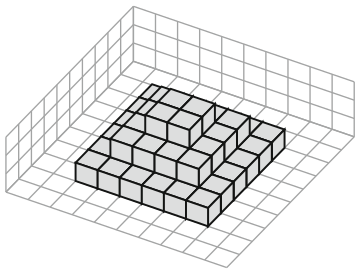}} \\ \cline{1-10} 
\multicolumn{1}{|c|}{Method} & A  & B  & \multicolumn{1}{c|}{C} & A & B & \multicolumn{1}{c|}{C} & A & B & \multicolumn{1}{c|}{C}     \\ \hline
\multicolumn{1}{|c|}{Sum of costs}           & 173                    & 176                    & \multicolumn{1}{c|}{179}   & 124                    & 128                    & \multicolumn{1}{c|}{128}   & -                      & {\color[HTML]{32CB00} 326}                    & \multicolumn{1}{c|}{326}   \\ \hline
\multicolumn{1}{|c|}{No. of timesteps}       & 13                     & 48                     & \multicolumn{1}{c|}{\color[HTML]{32CB00} 17}    & 13                     & 48                     & \multicolumn{1}{c|}{\color[HTML]{32CB00} 14}    & -                     & {\color[HTML]{32CB00} 106}                    & \multicolumn{1}{c|}{\color[HTML]{32CB00} 44}    \\ \hline
\multicolumn{1}{|c|}{Final Computation Time (in sec)} & 1030.3                 & {\color[HTML]{32CB00} 16.6}                   & \multicolumn{1}{c|}{11.1}  & 61.0                   & {\color[HTML]{32CB00} 11.5}                   & \multicolumn{1}{c|}{10.2}  & -                      & {\color[HTML]{32CB00} 49.6}                   & \multicolumn{1}{c|}{26.4}  \\ \hline
\multicolumn{1}{|c|}{Total Computation Time (in sec)} & 1115.0                 & {\color[HTML]{32CB00} 241.3}                  & \multicolumn{1}{c|}{259.4} & 139.0                  & {\color[HTML]{32CB00} 235.9}                  & \multicolumn{1}{c|}{198.1} & \textgreater{}10,000   & {\color[HTML]{32CB00} 377.2}                  & \multicolumn{1}{c|}{318.1} \\ \hline
                                             &                        & \multicolumn{1}{l}{}   & \multicolumn{1}{l}{}       & \multicolumn{1}{l}{}   & \multicolumn{1}{l}{}   & \multicolumn{1}{l}{}       &                        & \multicolumn{1}{l}{}   & \multicolumn{1}{l}{}       \\
\multicolumn{1}{l}{}                         &                        & \multicolumn{1}{l}{}   & \multicolumn{1}{l}{}       & \multicolumn{1}{l}{}   & \multicolumn{1}{l}{}   & \multicolumn{1}{l}{}       &                        & \multicolumn{1}{l}{}   & \multicolumn{1}{l}{}      
\end{tabular}
\end{table*}

\begin{table*}
\setlength\tabcolsep{10pt}
\centering
\begin{tabular}{cccccccccc}
\hline
\multicolumn{1}{|c|}{Test Structure} & \multicolumn{3}{c|}{\includegraphics[width=0.15\linewidth]{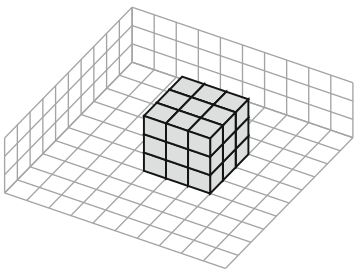}} & \multicolumn{3}{c|}{\includegraphics[width=0.15\linewidth]{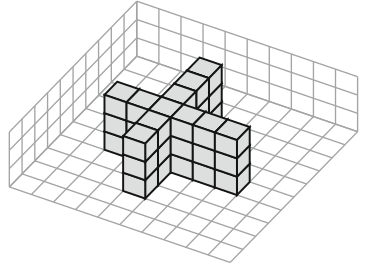}} & \multicolumn{3}{c|}{\includegraphics[width=0.15\linewidth]{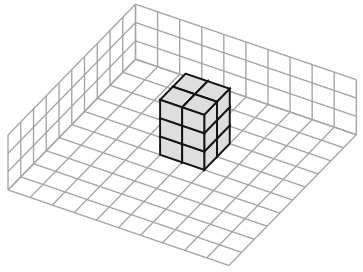}} \\  \cline{1-10} 
\multicolumn{1}{|c|}{Method} & A & B & \multicolumn{1}{c|}{C} & A & B & \multicolumn{1}{c|}{C} & A & B & \multicolumn{1}{c|}{C}     \\ \hline
\multicolumn{1}{|c|}{Sum of costs}           & -                      & {\color[HTML]{32CB00} 204}                    & \multicolumn{1}{c|}{263}   & -                      & {\color[HTML]{32CB00} 365}                    & \multicolumn{1}{c|}{381}   & 160                    & 153                    & \multicolumn{1}{c|}{161}   \\ \hline
\multicolumn{1}{|c|}{No. of timesteps}       & -                     & {\color[HTML]{32CB00} 113}                    & \multicolumn{1}{c|}{\color[HTML]{32CB00} 75}    & -                     & {\color[HTML]{32CB00} 130}                    & \multicolumn{1}{c|}{\color[HTML]{32CB00} 90}    & 21                     & 50                     & \multicolumn{1}{c|}{\color[HTML]{32CB00} 40}    \\ \hline
\multicolumn{1}{|c|}{Final Computation Time (in sec)} & -                      & {\color[HTML]{32CB00} 36.7}                   & \multicolumn{1}{c|}{610.7} & -                      & {\color[HTML]{32CB00} 37.1}                   & \multicolumn{1}{c|}{639.4} & 1215.3                 & {\color[HTML]{32CB00} 15.6}                   & \multicolumn{1}{c|}{265.9} \\ \hline
\multicolumn{1}{|c|}{Total Computation Time (in sec)} & \textgreater{}10,000   & {\color[HTML]{32CB00} 31.2}                   & \multicolumn{1}{c|}{758.6} & \textgreater{}10,000   & {\color[HTML]{32CB00} 27.8}                   & \multicolumn{1}{c|}{688.5} & 1715.2                 & {\color[HTML]{32CB00} 12.2}                   & \multicolumn{1}{c|}{287.9} \\ \hline
                                             &                        & \multicolumn{1}{l}{}   & \multicolumn{1}{l}{}       & \multicolumn{1}{l}{}   & \multicolumn{1}{l}{}   & \multicolumn{1}{l}{}       &                        & \multicolumn{1}{l}{}   & \multicolumn{1}{l}{}       \\
\multicolumn{1}{l}{}                         &                        & \multicolumn{1}{l}{}   & \multicolumn{1}{l}{}       & \multicolumn{1}{l}{}   & \multicolumn{1}{l}{}   & \multicolumn{1}{l}{}       &                        & \multicolumn{1}{l}{}   & \multicolumn{1}{l}{}      
\end{tabular}
\label{table:six_structures}
\end{table*}

\subsection{Comparison with Exact Approach on Random Structures}
We demonstrate the scope of our approach by testing all three optimization-based approaches on a set of randomly generated structures in two grid world sizes: 10x10x4 and 7x7x4. Structures were generated targeting ranges of occupancy percentage: the number of blocks in the input structure divided by the number of cells in the workspace. Out of the 100 test structures, half had an occupancy percentage between 40\% to 60\%, one-fourth had less than 40\%, and one-fourth had more than 60\%. The maximum number of robots permitted is 20 for the 10x10x4 environment and 6 for the 7x7x4 environment.


On average, for a set of test structures in a fixed environment size of 7X7X4, we noted that as the occupancy percentage increased from 30 to 70\%, the number of time steps increased from 76 to 122. This happens because the number of variables in the optimization model in the MILP formulation increases significantly, with the number of time steps following a linear trend. However, the solution computation time increases exponentially as the number of variables in the model increases. The exponential increase is illustrated by Fig. \ref{fig:solve time variables}. This is why decomposition is beneficial for run-time: solving several smaller MILP models (one for each substructure, smaller because substructures can usually be built in fewer time steps than the full structure) is much faster than solving one large model (for the entire structure at once).


For 100 random structures in the 10x10x4 grid world, we again tested our decomposition algorithm with MILP optimization. Table \ref{table:random_struct} shows the results for both environment sizes.

\begin{table}[h!]
\centering
\caption{Results for tests on random structures} 

Here, \textbf{EA - MILP-based Exact Approach, EAD - 3D decomposition with MILP-based Exact Approach}
\newline

\setlength\tabcolsep{2.5pt}
\begin{tabular}{ccccc}
\hline
\multicolumn{1}{|c|}{Environment Size}       & \multicolumn{2}{c|}{10x10x4}                       & \multicolumn{2}{c|}{7x7x4}                        \\ \hline
\multicolumn{1}{|c|}{Method}                 & A                    & \multicolumn{1}{c|}{B}      & A                    & \multicolumn{1}{c|}{B}     \\ \hline
\multicolumn{1}{|c|}{Sum of costs}           & -                    & \multicolumn{1}{c|}{384}    & 83.72                & \multicolumn{1}{c|}{97.12} \\ \cline{1-1}
\multicolumn{1}{|c|}{No. of timesteps}       & -                    & \multicolumn{1}{c|}{84}     & 18.01                & \multicolumn{1}{c|}{51.17} \\ \cline{1-1}
\multicolumn{1}{|c|}{Final Computation Time} & -                    & \multicolumn{1}{c|}{48.10}  & 229.41               & \multicolumn{1}{c|}{1.48}  \\ \cline{1-1}
\multicolumn{1}{|c|}{Total Computation Time} & \textgreater{}10,000 & \multicolumn{1}{c|}{567.99} & 423.51               & \multicolumn{1}{c|}{37.83} \\ \hline
                                             &                      & \multicolumn{1}{l}{}        & \multicolumn{1}{l}{} & \multicolumn{1}{l}{}       \\
\multicolumn{1}{l}{}                         &                      & \multicolumn{1}{l}{}        & \multicolumn{1}{l}{} & \multicolumn{1}{l}{}      
\end{tabular}
\label{table:random_struct}
\end{table}

\section{CONCLUSIONS}\label{conclusion}

In this paper, we presented an algorithm to decompose any input structure into substructures and obtain a feasible order to build the substructures. Experimental results showed that MILP optimization with decomposition has an order of magnitude improvement in the solution computation time compared to MILP without decomposition. However, the former demonstrated a higher makespan when the substructures were built sequentially or with basic parallelization.

Developing more sophisticated algorithms to construct substructures in parallel is a promising future direction. For example, one can determine the action sequence to construct each substructure that can be built in parallel. Then the action sequences can be post-processed to enforce collision and number of agents constraints. Further, the decomposition into substructures can be modified to optimize metrics like the number of scaffolding blocks required or the parallelism provided. 

Finally, the similarity between substructures can be leveraged to calculate the action sequence required to build the latest substructure using solutions of previous substructures. 

\end{document}